\pdfoutput=1

\documentclass[11pt]{article}

\usepackage[preprint]{acl}

\usepackage{times}
\usepackage{latexsym}
\usepackage{booktabs}

\usepackage[T1]{fontenc}

\usepackage[utf8]{inputenc}

\usepackage{microtype}

\usepackage{inconsolata}

\usepackage{graphicx}

\usepackage[english]{babel}
\usepackage[english=british]{csquotes}
\usepackage{amsmath}

\usepackage{tikz}
\usetikzlibrary{shapes.geometric, arrows.meta, positioning, decorations.pathreplacing}

\newcommand{\fone}{\(F_1\)}
\newcommand{\tildes}{\textasciitilde\textasciitilde\textasciitilde}

\title{Strategies for political-statement segmentation\\and labelling in unstructured text}

\author{Dmitry Nikolaev{\thanks{The authors contributed equally to this paper; the order is alphabetical.}} \\
  University of Manchester \\
  \texttt{dmitry.nikolaev@manchester.ac.uk} \\\And
  Sean Papay\footnotemark[1] \\
  University of Bamberg \\
  \texttt{sean.papay@uni-bamberg.de} \\}

\begin{document}
\maketitle
\begin{abstract}


Analysis of parliamentary speeches and political-party manifestos has become an integral area of 
computational study of political texts.
While speeches have been overwhelmingly analysed using unsupervised methods,
a large corpus of manifestos with by-statement political-stance labels
has been created by the participants of the MARPOR project. It has been recently shown
that these labels can be predicted by a neural model; 
however, the current approach relies on provided statement boundaries, limiting 
out-of-domain applicability. In this work, we propose and test a range of unified split-and-label 
frameworks---based on linear-chain CRFs, fine-tuned text-to-text models, and the combination of 
in-context learning with constrained decoding---that can be used to 
jointly segment and classify statements from raw textual data.
We show that our approaches achieve competitive accuracy when applied to raw text of political 
manifestos, and then demonstrate the research potential of our method by applying it to the
records of the UK House of Commons and tracing the political trajectories of four major parties in the 
last three decades.




\end{abstract}



\section{Introduction}

Among the genres used by politicians to communicate with each other and voters, two of the most important ones are 
party manifestos and speeches made in deliberative assemblies, such as the House of
Commons in the UK or the Bundestag in Germany. These sources are publicly available,
but their sheer volume makes manual analysis of them a very challenging task, and the study of
party manifestos and parliamentary debates has become one of the cornerstones of computational 
analysis of political texts
\citep[cf., among others,][]{parlaclarin-2022-parlaclarin,arroyo2022devolution,Muller_Proksch_2023}.

While members of the research community share an interest in analysing the stances expressed by politicians towards different
issues, the particular approaches taken for these two types of texts have largely differed.

The analysis of party manifestos has, to a large extent, coalesced around the labelling scheme developed
in the framework of the MARPOR project \citep{Volkens_2021} and used to manually annotate manifestos from more 
than 60 countries, written in almost 40 
languages.\footnote{\url{https://manifesto-project.wzb.eu/information/documents/corpus}}
MARPOR labels are attached to \textit{statements}, semantically coherent units on the sentence or sub-sentence level.
These labels correspond to political issues, such as national defence or migration, but often encompass both
an issue and a particular stance towards that issue. For example, label 504, \enquote{Welfare state expansion}, is assigned to 
\enquote{Favourable mentions of need to introduce, maintain or expand any public social service or social security scheme.}
Therefore, by means of simply counting different labels assigned to statements from a particular manifesto, it is possible to 
obtain a rather fine-grained representation of the political program expressed therein.

Until recently, efforts to assign these labels automatically had been largely unsuccessful and limited in 
scale \citep{dayanik-etal-2022-improving}.
It has been then shown by \citet{nikolaev-etal-2023-multilingual}\footnote{And concurrently, albeit in
a less rigorous fashion, by \citet{burst2023a,burst2023b}.} that contemporary multilingual
models can be used for adequate cross-lingual analyses. However, their approach relies on the availability of
statement boundaries, not provided by existing NLP tools, which limits the practical applicability of the
trained models.\footnote{It has been argued by \citet{daubler2012natural} that sentences, which NLP tools 
do aim to identify, are valid units of analysis in 
computational analyses of political texts.
The MARPOR annotation 
practices remain prevalent, however, and this is the setting we are targeting in this study.}

Conversely, the study of parliamentary debates, where labelled corpora are non-existent and the basic unit
is usually a whole speech, has overwhelmingly relied on unsupervised exploratory methods, such as topic
modelling, or even manual analysis, and targeted simple binary categories and aggregate scales
\citep{abercrombie2020sentiment,nanni2022political,skubic-fiser-2024-parliamentary}.\footnote{A~limited 
attempt at applying the MARPOR coding scheme to parliamentary data, again on the speech level, has been made 
by \citet{abercrombie-batista-navarro-2022-policy}, but it relies on a rather strong assumption that the whole 
speech revolves around the same narrow topic.}

The MARPOR categorization scheme has proven to be a powerful tool for political-text analysis,
applicable to almost any text in this domain,\footnote{Cf.\ an analysis of judges' decisions using this
framework by \citet{rosenthal2022judges}.} and the fact that labelled data and models trained on 
them only exist for party manifestos is largely a technical obstacle. Therefore, in this work we aim 
to solve the problem of projecting the MARPOR annotations to any running text.

In order to do this, we experiment with a series of models, spanning the landscape of Transformer-based
architectures. 

As a first step, we replace the encoder-based statement-level classifiers proposed by
\citet{nikolaev-etal-2023-multilingual} and \citet{burst2023a} with a linear-chain CRF layer 
\citep{laffertycrf} that learns to predict statement boundaries jointly with MARPOR labels using raw 
manifesto texts. This pipeline is very memory efficient and provides quick training and inference.
However, its ability to understand label sequences is limited by the expressive power of linear-chain 
CRFs, motivating investigation of autoregressive models.

As a more expressive but also more computationally demanding alternative, we propose using a pre-trained
T5-family model that is fine-tuned to split large textual chunks into statements and label these statements
at the same time.

Finally, we try to solve the task by using in-context learning, i.e.\ forgoing fine tuning and providing labelled
examples during inference with a state-of-the-art decoder-only 
model.\footnote{\citet{nikolaev-etal-2023-multilingual} showed that using long-input BERT-type model for 
directly predicting a scaling score, RILE, produced bad results, and it seems that language models is in 
general poorly suited for regression. Therefore in this study we only experiment with statement-level 
classification, which has additional practical benefits.}

We show that, even though fine-tuned T5-type models produce the best in-domain results, their high computational demands and slow inference limit their practical applicability to large-scale out-of-domain experiments.
For such cases, the CRF-based model makes for a better choice, showing a slight performance degradation on in-domain evaluation but orders-of-magnitude faster inference.

Equipped with our CRF model, capable of efficiently segmenting and labelling statements from raw text,
we perform an exploratory analysis of the UK parliamentary records.\footnote{A secondary study of
Australian data is reported in the Appendix.}
We further discuss the problem of the parliamentary data being  out-of-domain, especially in terms of label 
sequences, and propose to mitigate it using model ensembling.

\section{Data}

We target the same original-language and translated subsets of the MARPOR dataset as used by 
\citet{nikolaev-etal-2023-multilingual}.\footnote{Available at \url{https://osf.io/aypxd/}}
Out of the two settings explored in their paper, leave-one-country-out
and old-vs.-new, we adopted the former as it is more challenging.

Since training and testing larger
models on all 41 countries from their dataset is not practicable, we adopted the following approach:
after a complete preliminary analysis done using the XML-R + CRF approach, we split the countries into quartiles
based on the test-set performance. We then selected a country from the middle of the each quartile and used this
country's manifestos as a test set for subsequent experiments. For each of the test countries we also used the 
same set of dev-set-countries' manifestos when training the CRF and fine-tuning Flan T5.\footnote{The test 
countries with their respective dev-set countries are as follows: Denmark (Netherlands, Turkey),
Netherlands (Mexico, Slovakia), Bulgaria (Chile, Georgia), Uruguay (Austria, Czech Republic).}

The dataset for the exploratory analysis of parliamentary records is described in \S~\ref{sec:analysis}.

\section{Methods}
The problem of jointly segmenting and classifying statements from text is an
example of a span identification, or extraction, task.
In spite of the fact that all models we use rely on the same underlying Transformer architecture, 
they demand different approaches to task operationalisation and input/output encoding.
We specify them below.\footnote{The training code for the study 
will be uploaded to a public repository in case of acceptance.}



\subsection{CRF}

\paragraph{Input formatting.}

Following standard practice, we encode the statements using the BIO scheme
\cite{ramshaw-marcus-1995-text} and use a sequence-labelling model to predict token-wise labels.
As the spans we are extracting form a total cover of our texts,
the \textsc{O} label is ultimately only used for padding and BOS/EOS tokens.

\paragraph{The architecture.}

Our model combines a linear-chain CRF with a pre-trained XLM-RoBERTa (XLM-R) encoder
\cite{DBLP:journals/corr/abs-1911-02116} providing token-wise emission scores for the CRF.
Due to XLM-R's multi-lingual pre-training, we are able to directly
use manifestos in their original languages.


As the political manifestos we train on are significantly larger than our encoder's context window, 
we divide the input text into multiple overlapping windows, feed these windows to our encoder independently, 
and stitch together the contextualized representations obtained from the centre of each window for use as input to the CRF.
In this way, we can process sequences of arbitrary length, while still ensuring that each token's representation 
was generated with adequate context to both left and right.

During inference, we feed entire documents as input to our model in this manner, irrespective of length, while during training, for performance reasons, we limit model inputs to 1024 tokens, yielding a maximum of four overlapping windows.
A~complete description of our model, including hyperparameters and splitting procedures, is provided in Appendix~\ref{app:hyperparameters}.

\paragraph{Training.}
For each cross-validation split, we initialize our encoder with pre-trained XLM-R weights 
and randomly initialize all other model weights.
We jointly optimize all model weights on negative-log-likelihood loss using mini-batch gradient descent.
During training, we periodically calculate the model's $F_1$-score on the held-out development set in order 
to guide early stopping. After twenty such evaluations with no improvement, we terminate training, retaining 
model weights from the training step that yielded the highest dev-set $F_1$-score.

\subsection{Fine-tuned Flan-T5}

We use the pre-trained version of Flan T5 XL from HuggingFace\footnote{\url{https://huggingface.co/google/flan-t5-xl}}
as the base model. Since Flan T5 is English only, we use the translated version of the dataset.

\paragraph{Input formatting.}

Thanks to using relative attention Flan T5 is able to handle contexts of arbitrary length. However,
due to high memory constraints we split the MARPOR manifestos input into chunks of 260 tokens,
as defined by the model's tokeniser.
Input consisted of raw text, and the output consisted of input statements followed by their MARPOR
label followed by a triple 
tilde.\footnote{Originally we experimented with splitting labelled statements with line-breaks,
but line-breaks were replaced by single spaces during decoding. The fine-tuned model also refused to
reconstruct triple tildes, but it consistently replaced them with <unk>, which we then used to extract
statements.} A~sample input-output pair is shown in Appendix~\ref{app:sec:input-output}.

\paragraph{Training.}

The model was trained with the standard cross-entropy loss
using the AdamW optimiser \citep{loshchilov2019adamw} with the learning rate of \(10^{-5}\) for 5 epochs,
and we selected the checkpoint that performed best on the dev set for
testing.\footnote{We used the same cross-entropy loss to select the checkpoint and not span-extraction
and label-prediction accuracy. The latter would be beneficial, but inference with T5 XL is very slow,
so we only used it for the test set.} We then decoded greedily at test time.

\subsection{In-context learning with Llama 3.1}
Our final model is an in-context learning approach utilizing Llama 3.1 8B Instruct \citep{dubey2024llama}, an instruction-tuned large language model.
We use the provided model weights as-is and do not further fine tune this model.
As Llama 3.1 does not support the vast majority of languages present in the MARPOR corpus, 
we again use English-language translations of the manifestos.

In order to obtain useful predictions from this pre-trained model, we leverage few-shot in-context learning \citep{NEURIPS2020_1457c0d6, wei2022emergent} with decoding-time constraints.

We present the model with a short English-language system message, tasking it with segmenting and classifying claims from a provided snippet from a party manifesto.
We then present a fabricated chat history of thirty task-response pairs.
For each of these, a user message presents a snippet of a party manifesto, and an agent message
parrots the same text back, inserting statement labels after each statement.
These in-context learning examples are drawn uniformly randomly from the training partition, agent responses reflecting gold-standard segmentations and labellings.
Statements are labelled by the English name of their category titles, parenthesized \mbox{[(like this)]}.   
By using descriptive English names, as opposed to numeric IDs, the model can leverage existing semantic knowledge obtained from its pre-training when assigning labels to statements.

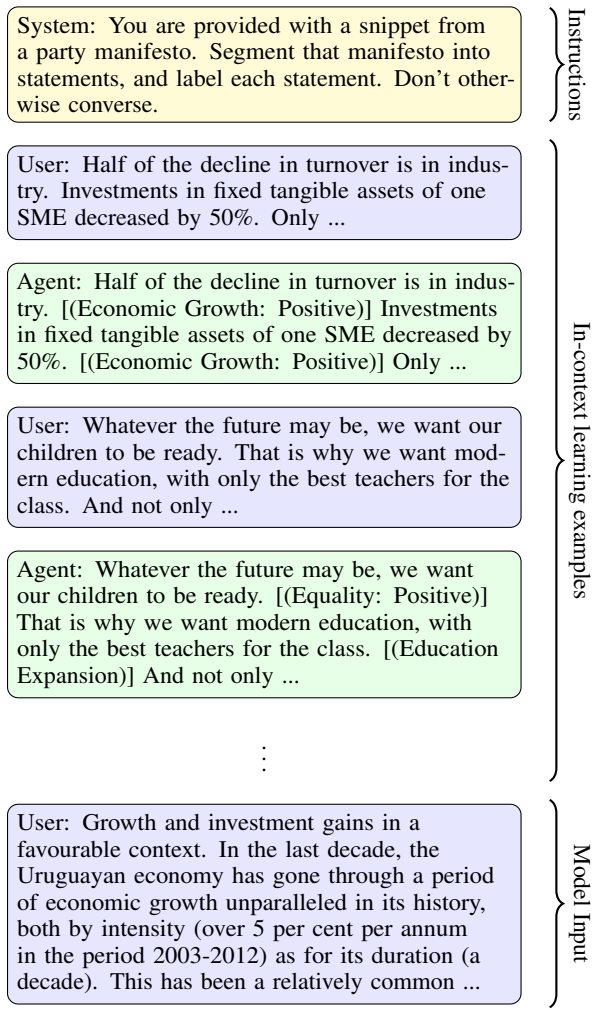
\begin{figure}[t]
    \centering
\begin{tikzpicture}[
    font=\footnotesize, 
    user/.style={draw, fill=blue!10, rounded corners, text width=6.5cm, align=left, minimum height=0.5cm, anchor=west},
    agent/.style={draw, fill=green!10, rounded corners, text width=6.5cm, align=left, minimum height=0.5cm, anchor=west},
    system/.style={draw, fill=yellow!20, rounded corners, text width=6.5cm, align=left, minimum height=0.5cm, anchor=west},
    ellipsis/.style={align=center, font=\Huge}, 
    every node/.style={anchor=west},
    node distance=0.3cm 
]

\node (sys) [system] {System: You are provided with a snippet from a party manifesto. Segment that manifesto into statements, and label each statement. Don't otherwise converse.};

\node (user1) [user, below=of sys] {User: Half of the decline in turnover is in industry. Investments in fixed tangible assets of one SME decreased by 50\%. Only ...};
\node (agent1) [agent, below=of user1] {Agent: Half of the decline in turnover is in industry. [(Economic Growth: Positive)] Investments in fixed tangible assets of one SME decreased by 50\%. [(Economic Growth: Positive)] Only ...};

\node (user2) [user, below=of agent1] {User: Whatever the future may be, we want our children to be ready. That is why we want modern education, with only the best teachers for the class. And not only ...};
\node (agent2) [agent, below=of user2] {Agent: Whatever the future may be, we want our children to be ready. [(Equality: Positive)] That is why we want modern education, with only the best teachers for the class. [(Education Expansion)] And not only ...};

\node (ellipsis) [below=of agent2] {$\vdots$};

\node (final_user) [user, below=of ellipsis] {User: Growth and investment gains in a favourable context. In the last decade, the Uruguayan economy has gone through a period of economic growth unparalleled in its history, both by intensity (over 5 per cent per annum in the period 2003-2012)  as for its duration (a decade). This has been a relatively common ...};


\draw [decorate,decoration={brace,amplitude=5pt,raise=4pt}, thick] (7, 0.75) -- (7,-.75) node[midway,rotate=270,yshift=.25cm,anchor=south] {Instructions};

\draw [decorate,decoration={brace,amplitude=5pt,raise=4pt}, thick] (7,-1) -- (7,-9.5) node[midway,rotate=270,yshift=.25cm,anchor=south] {In-context learning examples};

\draw [decorate,decoration={brace,amplitude=5pt,raise=4pt}, thick] (7,-9.75) -- (7,-12.5) node[midway,rotate=270,yshift=.25cm,anchor=south] {Model Input};

\end{tikzpicture}
    \caption{An example of an in-context learning prompt, comprising natural-language instructions, in-context learning examples, and the input text. The instructions are shown verbatim; in-context learning examples shown are real examples from the dataset but are truncated for space. The model's response to this prompt, decoded with constraints, will constitute the prediction for the input text.}
    \label{fig:icl}
\end{figure}

After these 30 in-context learning examples, we present a final user message, this time presenting a snippet of a manifesto taken from the test partition.
At this point, the model is left to generate a continuation response labeling and segmenting this snippet.
Figure~\ref{fig:icl} illustrates a prompt built in this way.

As we are specifically interested in statement segmentations and labellings, and not in a conversational 
response the model was instruction-tuned to provide, we make use of decoding-time constraints to severely limit 
the possible output space. At each time-step, we only consider possible continuations that either (i) parrot the 
next token as was present in the input snippet, (ii) begin a statement label tag, or (iii) continue a 
previously-begun statement label tag in a way that can lead to a legal tag with a valid statement name.
A token trie is used to efficiently track legal continuations for already-begun tags.

In this way, every allowed response corresponds one-to-one with a possible segmentation and labelling of the 
input sequence. As greedy decoding might lead the model to commit to tokens with no legal high-probability 
continuations, we decode from this constrained model with beam search of beam width three.

\subsection{Evaluation}

We evaluate the models on two tasks: (i) statement segmentation and MARPOR-label prediction and (ii)~a 
\enquote{downstream} task of political scaling, i.e.\ assigning to political texts numerical scores that 
characterise their position on a certain continuum. We use \fone-scores for (i) and target the Standard 
Right--Left Scale, a.k.a.\ the RILE score, as the most commonly used political scale. It is computed using 
the following formula:
\begin{equation}
    \text{RILE} = \dfrac{R - L}{R + L + O}
    \label{eq:rile}
\end{equation}
\textit{R} and \textit{L} stand for the number of right- and left-leaning statements in the target manifesto,
respectively, and \textit{O} stands for other statements. The categories making up the \textit{R} and 
\textit{L} groupings are shown in Table~\ref{tab:rile-categories} in the Appendix. 
See \citet{volkens2013rile} for more details.

\section{Results}

\begin{table}[t]
\small
\centering
\begin{tabular}{lrrrrr}
\toprule
Model & Precision & Recall & $F_1$ & RILE\\
\midrule
CRF & 41.3 & 40.7 & 40.7 & 0.74\\
CRF+Oracle & 48.0  & 50.0 & 48.6 & 0.75\\
\midrule
Baseline (XLM-R) & -- & -- & 44 & 0.73\\
Baseline (MT) & -- & -- & 44 & 0.71\\
\bottomrule
\end{tabular}
\caption{The results of predicting MARPOR labels and RILE scores for held-out manifestos.
Precision, recall, and $F_1$-scores are 
weighted by support in the true labels.
Performance on RILE is measured as Spearman correlation of computed and gold scores.
MT denotes using an English SBERT encoder with translated inputs.}
\label{tab:f1}
\end{table}

\begin{table}[t]
\centering
{\small
\begin{tabular}{@{}lllll@{}}
\toprule
           & Denmark & Netherlands & Bulgaria & Uruguay \\ \midrule
CRF        & 45.4    & 42.33       & \textbf{41.2}    & 33      \\
Flan       & \textbf{48.3}    & \textbf{43.5}        & 40.1     &    \textbf{37.5}     \\
Flan+      &   40      &   37.7          &   37.9       &    31.1     \\
ICL        & 32.7   &      31.3       & 24.9     & 25.1   \\ \midrule
CRF        & 40.72   & 40.95       & 38.72    & 24.96   \\
Flan       & \textbf{45.52}   & \textbf{43.3}        & \textbf{42.7}     & \textbf{34.3}    \\
Flan+ & 39.4    & 37.2        & 39.6     & 28.4    \\
ICL        & 29.34   &     30.89        & 26.88    & 22.97   \\ \bottomrule
\end{tabular}}
\caption{\fone-scores for extracted and labelled spans in the test sets.
Micro-averaged scores are in the upper part of the table, and the lower
part presents scores averaged by manifesto.
Flan+ stands for combining span extraction using
Flan T5 XL with label assignment using \citeauthor{nikolaev-etal-2023-multilingual}'s
SBERT-based model. ICL is Llama 3.1 8B Instruct.}
\label{tab:f1-micro}
\end{table}

\begin{table}[t]
\centering
{\small
\begin{tabular}{@{}lllll@{}}
\toprule
           & Denmark & Netherlands & Bulgaria & Uruguay \\ \midrule
CRF        & 0.67    & 0.79       & 0.54    & 0.9      \\
Flan       & \textbf{0.84}   &   \textbf{0.9}          &     0.45     &     1    \\
Flan+      &  0.78       &      0.86       &    0.59      &    1     \\
ICL        & 0.71   &     0.78        & \textbf{0.62}     & 1   \\ \bottomrule
\end{tabular}}
\caption{RILE scores computed using predicted labels.
See the caption of Table~\ref{tab:f1-micro} for model abbreviations.
}
\label{tab:rile}
\end{table}

With the leave-one-country-out cross-validation setting, we obtain one set of
model predictions for one test country. For the CRF-based model, where all 41 countries
were processed, we analyse our results on the union of by-country predictions.
For Flan T5 XL and Llama, we report the results for each test country individually.


\subsection{CRF-based segmentation}

Table~\ref{tab:f1} summarises the performance of the CRF-based model in terms of macro-averaged $F_1$-scores
for exact span-and-label matches, weighted by class frequency, and compares its results with those from
\citet{nikolaev-etal-2023-multilingual} where, as in all other prior work, gold statement boundaries 
were assumed.

We find that after replacing gold statement boundaries and a unigram-based classifier by an end-to-end
CRF model we obtain the results that differ by less than four percentage points.
We can interpret numerical differences in $F_1$-scores as the result of two factors: differences in the two models' 
competency at \textit{classifying} claims, and additional challenges introduced by the task of determining claim 
\textit{boundaries}, which are only faced by our model.

We can attempt to disentangle these two factors by providing our model with an oracle
for span boundaries. This can be accomplished at decoding time by
constraining \cite{papay2022constraining} our CRF output as follows: our CRF \textit{must} output some 
begin tag wherever the true label sequence has a begin tag, and it \textit{must not} output a begin tag 
wherever the true label sequence does not have a begin tag.
In this way, we can ensure that our model's statement boundaries match the true boundaries, 
while still allowing our CRF to choose which MARPOR category to assign to each statement.

Under this setting, we find that our model actually outperforms the classifier-based baseline by more 
than 4 percentage points. As both models use XLM-R as an encoder, we cannot ascribe this performance 
difference to quality of latent representations. Instead, we suspect that our CRF-based model's ability
to model interactions between adjacent statement labels gives it an edge against the classifier-based baseline,
which must predict statement labels independently.

Interestingly, even though our oracle-free model loses to the baselines on $F_1$, it still leads to better 
estimates of manifesto-level RILE scores, 
which was the main target for \citet{nikolaev-etal-2023-multilingual}.
Mistakes made by the new model therefore seem to be less \enquote{damaging} 
in the sense that, e.g., left-leaning stances are not identified as neutral or right-leaning.

\subsection{Text-to-text and in-context learning}

The analysis above highlights the importance of incorporating sequential information in political-stement
labelling. Given that the CRF is hamstrung by its inability to model non-immediate context, we can expect 
autoregressive models attending to long histories to outperform it. Large language models with decoders are a natural fit for this task. 

Further, adding constraints on the decoding or an explicit copy mechanism is a natural way
of simplifying the task of regenerating the input, and we did add constrained decoding to Llama 3.1.
Preliminary tests of fine-tuned Flan T5 XL, however, showed that the model very rarely garbles the input, 
so in the interest of simplicity and decoding speed (see \S~\ref{sec:compute})
we resorted to the greedy strategy.

The results for span extraction and labelling are shown in Table~\ref{tab:f1-micro}. With extracted spans,
gold-label-weighted \fone\, becomes less interpretable, and we revert to simple micro-averaging and macro-averaging
across manifestos. The correlations between RILE scores computed using predicted
and gold labels for all models are shown in Table~\ref{tab:rile}. The test countries can be roughly split in 
three groups in terms of model performance. 

The first group
consists of Denmark and Netherlands. Both these countries have large test sets, with manifestos written
in comparatively well-resourced Western European languages. This ensures higher quality of both multilingual
embeddings (used by the CRF model) and the MT models, which provide inputs to LLMs. In both cases, we
see the same outcome: the vanilla Flan T5 XL is a clear winner in terms of classification accuracy, with the CRF
model a more or less close second. 

In terms of downstream RILE scores, Flan T5 is again the best model, but the second
place is now taken by the combination of Flan-derived spans with SBERT-assigned labels, and the CRF model
loses even to the Llama-based model, whose accuracy is very low. This further reinforces the conclusions 
by \citet{nikolaev-etal-2023-multilingual} that when it comes to computing RILE scores, the nature of the
errors made by a given models 
becomes more important that its actual accuracy.

The second group consists of Uruguay, which is a very hard label-prediction task (Flan T5 attains an \fone-score
of 37.5, and all others do even worse), but a much easier scaling task, with correlations everywhere close to
1. The latter result, however, should be taken with a grain of salt since the test-set size is small (4 manifestos).

Finally, the most complicated case is presented by Bulgaria, which is closer to Uruguay in terms of span and 
label accuracy, with a minimal difference between CRF and Flan T5 in terms of the \fone-score, but where the
best performance on RILE is attained by the Llama-based setup. Most intriguingly, the performance of Flan T5 on
the RILE task is the worst among all the models.

If we regard Bulgaria as a sort of outlier with high-variance results induced by lower-quality embeddings or
translations, we may tentatively conclude that

\begin{enumerate}
    \item Using a fine-tuned Transformer-based model for span extraction and labelling provides a modest boost in
    performance over the CRF-based approach, even without constrained decoding.
    \item Conversely, using constrained decoding for multi-label classification in the in-context-learning
    setting does not yet lead to good results. This may be overcome by resorting to larger models or
    longer contexts; however, see \S~\ref{sec:compute} below.
    \item In contrast to exact label prediction, RILE-based scaling seems to be an easy task, with even
    constrained Llama 3.1 providing results on par with those reported by 
    \citet{nikolaev-etal-2023-multilingual}.
    This suggests that for coarse-grained analysis bypassing fine-tuning is already a valid strategy.
\end{enumerate}

\section{Discussion of computational demands}
\label{sec:compute}

In this section, we contrast computational demands of different approaches. We show that while training demands
of even bigger models that we use are manageable, given access to typical research-grade infrastructure,
inference on them becomes limited to hundreds, at most thousands of examples, which limits their
applicability to larger corpora in computational political science numbering millions of data points.

\subsection{Training}

\textbf{CRF + XLM-R} has relatively low demands for training, particularly when taking into account its much lower memory footprint than most modern autoregressive models: training required 6.87 GiB of GPU memory, and up to six independent models could be trained simultaneously on a single NVIDIA RTX A6000 GPU.
In this parallel training regime, each training process took about 1.08 seconds to complete a single training step with a batch size of one.

Fine-tuning \textbf{Flan T5 XL} is moderately demanding: while training on four NVIDIA A100 40 gigabyte GPUs, 
one batch of two 260-token inputs takes approximately 1.3 seconds for a forward and a backward pass.
While this is comparable to the CRF, fine-tuning Flan T5 XL requires approximately 60 gigabytes of GPU memory, limiting the ability to perform such fine-tuning on lower-end hardware and precluding the parallelisation of multiple training runs as was possible for the CRF-based model.

The in-context-learning setup does not demand a training stage.


\subsection{Inference}
Not relying on autoregression and benefiting from a smaller model size, inference was quite fast with the \textbf{CRF} model, averaging just over 3000 tokens per second.\footnote{For comparability, all inference 
speeds are reported in terms of Flan T5 XL tokenisation.}
Furthermore, as was the case with training, the model's small memory footprint allowed multiple 
inference procedures to be parallelised on a single GPU.

With sequential decoding in inference, the time demands of the two autoregressive models are almost prohibitive: \textbf{Flan T5 XL} performed inference at a rate of 26 tokens per second, and \textbf{Llama 3.1 8B}, requiring a long context for in-context-learning examples and beam-search decoding, averaged just under 3 tokens per second.
Such slow inference time makes these models infeasible to apply to large corpora such as UK or Australian Hansard for targeted experiments.

\section{Analysis of parliamentary debates}
\label{sec:analysis}

We now turn to the analysis of parliamentary data to show how our raw-text-capable
CRF-based model can be applied in another domain. While it is likely less powerful than
fine-tuned Flan T5 XL, it is incomparably faster in inference and can be used to process large
corpora without access to massive computational resources.

\subsection{Preliminaries}
\label{ssec:analysis-intro}

We apply our model to the records of parliamentary debates published as so-called Hansards
in the UK and some of the Commonwealth countries. Our primary data come from the UK version of
Hansard,\footnote{\url{https://hansard.parliament.uk/}} with a similar analysis for Australia presented in
Appendix~\ref{sec:hansard-au-analysis}. There is no published dataset of UK parliamentary debates annotated with CMP
labels.\footnote{\citet{abercrombie-batista-navarro-2022-policy} 
assigned CMP labels to a set of \textit{motions}, i.e.\ 
statements calling for a vote on a bill, and used these as gold labels for speeches responding to this motion.
The choice of the label, however, depends on the contents of the bill and not on the text of the motion itself.}
Therefore our analysis is exploratory, and it may be validated by evaluating 
the reasonableness and insightfulness of the revealed trends.

Preliminary analysis of the labels assigned by the CRF model demonstrated that, apart from the core of 
semantically relevant statements, it often assigned more general or technical statements that MARPOR labels 
as \enquote{Other} to other classes, most likely because topic sequences in the manifesto data differ 
significantly from those in parliamentary speeches. In order to mitigate this issue, we resorted to conservative 
model ensembling, and only included in the analysis statements on which our model and the classifier by 
\citet{nikolaev-etal-2023-multilingual}---with statement boundaries provided by the CRF model---agreed.
This happened in 38\% of cases (39.6\% on the Australian Hansard), which gives around 7 million 
statements for analysis. A randomised manual inspection of statements given different labels
(see examples in Appendix~\ref{sec:statement-examples}) showed that the performance of the ensemble model is 
good both in terms of statement boundaries and assigned labels. The only problematic category is 305, 
\enquote{Political authority}, which seems to lack a coherent core in the source data and competes with 
\enquote{Other} for general or procedural statements.

For the sake of robustness, we further restrict ourselves to statements made by members of
four major parties, the Conservative Party, the Labour Party, the Liberal Democrats (LibDems), and the Scottish National 
Party (SNP), between 1990 and 2019.
As Figure~\ref{fig:hansard-uk-statements} in Appendix~\ref{sec:stats} shows, the number of statements made by each party is
roughly proportional to its success in the preceding elections, with Conservatives and Labour dominating throughout
and SNP overtaking LibDems after 2015.

\subsection{Party trajectories}
\label{ssec:trajectories}

\begin{figure}[t]
    \centering
    \includegraphics[width=0.95\linewidth]{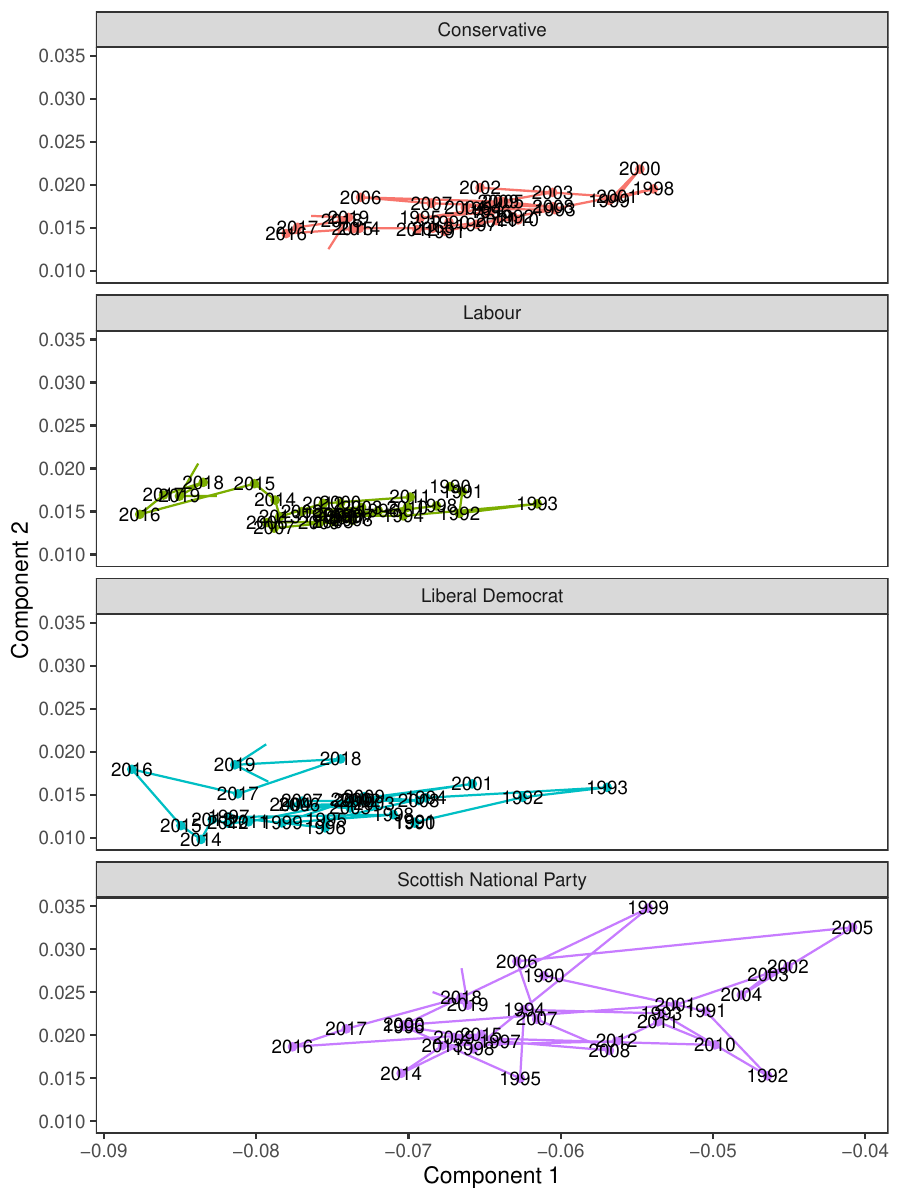}
    \caption{Political trajectories of major UK parties traced by projecting yearly salience vectors of CMP categories in their parliamentary speeches using non-negative matrix factorization and the original CMP data as the training set.}
    \label{fig:nmf-trajectories}
\end{figure}

In order to trace political evolution of major UK parties as reflected by statements their members made in 
the House of Commons, we use path diagrams. Each data point represents a distribution of 
CMP labels attached to statements made by a party in a given year. 
To derive the axes, we use non-negative matrix factorization with 2 components\footnote{Implemented 
in \href{https://scikit-learn.org/stable/modules/generated/sklearn.decomposition.NMF.html}{scikit-learn}.}
trained on the original CMP data with label counts aggregated by manifesto. This provides us with a
\enquote{universal salience baseline}.
We then use the trained model to project UK parliamentary data on the same axes.

The results of this procedure are shown in Figure~\ref{fig:nmf-trajectories}. Similar to previous work
on political-text scaling \citep{rheault2020word,ceron-etal-2023-additive}, an axis emerges that can be 
understood as politically left-vs.-right.
In our case, the first component 
portrays both Conservatives and Labour as largely centrist parties, 
with Labour spending several years (2015--2019) as a more left-wing one. This ties in nicely with the fact that
in 2015--2020 the party was lead by Jeremy Corbyn, who was noted for leading the party towards the radical left
\citep{goodger2022corbyn}.
%
LibDems, a centre-left party, is also to the left of Conservatives, while SNP, 
social democratic in terms of its social and economic policies but with a distinct nationalistic agenda 
\citep{mitchell2011snp}, shown as the most right-wing one.\footnote{The more traditional way of placing each of the parties
on the right--left scale using the MARPOR RILE formula \citep{volkens2013rile} is shown in Appendix~\ref{sec:uk-riles}.}

Our analysis can be contrasted by that by \citet[12]{rheault2020word}, who used averaged word embeddings. They portray
Labour as strictly to the left of Conservatives at all times with LibDems always occupying middle ground. Given the 
amount of convergence and shared values, e.g.\ on the expansion of welfare state, among the British political 
parties \citep{quinn2008,goodger2022corbyn}, this picture seems too simplistic.

\section{Related work}

As far as we are aware, no prior work addresses the problem of assigning MARPOR labels to
raw text, and the efforts were focused on providing higher-level stance or scaling analyses.
For example, \citet{subramanian-etal-2018-hierarchical} provided manifesto-level scaling scores by 
aggregating over LSTM-based representations of 
sentences and taking into account historical RILE values,
while \citet{liu-etal-2022-politics} present a model for determining ideology and stance, where both
target values are encoded as binary or 3-element scales.

The problem of automatically assigning contentful
MARPOR labels to statements in party manifestos was first addressed on a smaller scale by
\citet{dayanik-etal-2022-improving} and \citet{ceron-etal-2023-additive}, and then in a larger
cross-lingual setting by \citet{nikolaev-etal-2023-multilingual} and \citet{burst2023a,burst2023b}.
All these studies assumed, however, that gold statement boundaries are provided, which contrasts
with the fact that many MARPOR statements consist of sub-sentences, which demands a dedicated
span-extraction module.

The necessity of completely splitting the input into sub-sentence-level chunks contrasts our setting
with span-extraction tasks, such as NER, and more straightforward
sentence-segmentation settings, where the need for domain-specific approaches has also been recognised.
In the latter area, CRF- and encoder-based approaches continue to demonstrate
strong results, cf.\ \citet{Brugger2023MultiLegalSBDAM} for a domain-specific example and 
\citet{frohmann2024segment} for a general model. In a manner similar to ours,
\citet{mccarthy-etal-2023-long} contrast CRF-based approaches to text segmentation to using
LLMs with constrained decoding.

\section{Conclusion}
\label{sec:conclusion}

The analysis of political texts has long been impeded by the absence of a model providing 
identification and fine-grained semantic labelling of statements. In this work, we show that it 
is possible to assign statement boundaries and stance labels at the same time.
Using well-proven methods, a BERT-type encoder with a CRF layer, we reach good performance on the 
manifesto data and then demonstrate that our model can provide insightful analyses of 
parliamentary data in the standard MARPOR framework. We furthermore show that better results can
potentially be attained using simple fine-tuning of a large text-to-text model, but
its low inference speed precludes its use for large-scale exploratory studies.
Finding ways of accelerating inference on high-volume raw-text segmentation and analysis 
is an important avenue for future work.

\section*{Limitations}
For in-domain performance, the breadth of languages covered made an in-depth qualitative analysis impossible, 
as the majority of manifestos were written in languages not spoken by the authors.
For the autoregressive models, computational costs prevented us from performing a full-scale comparison against the CRF across all 41 countries.
Due to a lack of labeled data for the parlimentary debates domain, we were unable to quantitatively evaluate 
our models' out-of-domain performance.
Furthermore, our exploratory analysis of parlimentary debates was limited to two English-speaking countries.

\bibliography{main}

\appendix

\section*{Appendix}

\section{RILE categories}

\begin{table*}[t]
\centering
\small
\begin{tabular}{ll}
Right emphasis &
  \begin{tabular}[c]{@{}l@{}}Military: Positive, Freedom, Human Rights, Constitutionalism: Positive, \\ 
  Political Authority, Free Enterprise, Economic Incentives, Protectionism: \\ Negative, Economic Orthodoxy, 
  Social Services Limitation, National Way \\ of Life: Positive, Traditional Morality: Positive, Law and Order, 
  Social Harmony\end{tabular} \\ \cline{2-2} 
Left emphasis &
  \begin{tabular}[c]{@{}l@{}}Decolonisation, Anti-imperialism, Military: Negative, Peace, Internationalism: \\ 
  Positive, Democracy, Regulate Capitalism, Market, Economic Planning, \\ Protectionism: Positive, Controlled Economy, Nationalisation, Social Services: \\ Expansion, Education: Expansion, Labour Groups: Positive\end{tabular} \\ 
\end{tabular}
\caption{The MARPOR categories used for calculating the RILE score.}\label{tab:rile-categories}
\end{table*}

MARPOR categories used for computing the RILE score are shown in Table~\ref{tab:rile-categories}.

\section{Model details}
\label{app:hyperparameters}
This appendix details the specifics of our CRF-based-model architecture and training procedure.

\subsection{Model architecture}
As an encoder, we used the XLM-RoBERTa \cite{DBLP:journals/corr/abs-1911-02116} pretrained model, with weights obtained from
HuggingFace.
As almost all inputs exceeded the 512-token context length of this model, we adopted an overlapping-window approach to encoding longer sequences.

After tokenizing documents in their entirety, we define a number of overlapping 512-token windows to use as independent inputs to our encoder.
A new window starts every 256 tokens, such that, except for the start and end of the text, each token is part of exactly two windows.
These windows are all used as independent inputs to XLM-RoBERTa, yielding two separate representations for each interior token (one for each window that token is a part of).
We take the embeddings from the central half of each window (tokens indexed 64 to 192) and concatenate these to form our input representations -- this results in exactly one contextualized vector for each input token and always ensures that these vectors are calculated with adequate left- and right-context.

Our BIO labeling scheme leaves us with 275 labels.
We pass our input representations through a 275-unit linear layer in order to obtain emission scores for our CRF.
Transition scores are stored explicitly in a $275\times275$ weight matrix, which is initialized randomly.

\subsection{Training}
We optimize all parameters jointly, fine-tuning the XLM-RoBERTa weights while learning weights for our linear layer and transition matrix.
We utilize the Adam optimizer \cite{2015-kingma} with an initial learning weight of $5\times10^{-6}$.
Due to the length of our documents, we use a batch size of 1.
We further limit documents to a length of 1024 during training.

Every 2000 training steps, we evaluate model $F_1$-score on the held-out development set in order to guide early stopping.
After twenty such evaluations with no improvement, we terminate training, retaining model weights from the training step which yielded the highest in dev-set $F_1$-score.

The 41 splits were trained in parallel across a number of NVIDIA GeForce GTX 1080 Ti and NVIDIA RTX A6000 GPUs.
Training time varied significantly per split but was generally on the order of about 4 days.

\section{Flan T5 XL input-output example}
\label{app:sec:input-output}

\paragraph{Input:} \texttt{\small We want to renovate the quality society in the current quantity society. In a quality society, we do not always have to do more, but better. Quality means services and sustainable products that do not waste energy and natural resources. In working life, quality means fairness, equality and enjoyment – and that everyone can participate. The quality of life is improved by having more time for family and loved ones. We want stone-foot shops, domestic services, village economy promote the export of environmental technology and computer games. We want the goods purchased not only to be able to carry home but also to be used, and we want them to be repaired or repaired by professionals. Work must continue to be done at hand and without it, because a green quality society is a society for everyone. The aim of green policy is to restore a more gentle, equal and environmentally friendly society. From single use to sustainability: products, goods, nature and climate must last from generation to generation. We want an eco-efficient economy based on services, knowledge and knowledge, where healthier and more long-term people can influence their own lives. We want a more diverse nature, a more nuanced environment and a richer cultural environment. We want a world where children do well to live.}

\paragraph{Output:} \texttt{\small We want to renovate the quality society in the current quantity society. 416\tildes In a quality society, we do not always have to do more, but better. 416\tildes Quality means services 504\tildes and sustainable products that do not waste energy and natural resources. 416\tildes In working life, quality means fairness, equality and enjoyment – and that everyone can participate. 701\tildes The quality of life is improved by having more time for family and loved ones. 603\tildes We want stone-foot shops, domestic services, village economy 408\tildes promote the export of environmental technology and computer games. 410\tildes We want the goods purchased not only to be able to carry home but also to be used, and we want them to be repaired or repaired by professionals. 416\tildes Work must continue to be done at hand and without it, because a green quality society is a society for everyone. 701\tildes The aim of green policy is to restore a more gentle, equal and environmentally friendly society. 416\tildes From single use to sustainability: products, goods, nature and climate must last from generation to generation. 416\tildes We want an eco-efficient economy based on services, knowledge and knowledge, where healthier and more long-term people can influence their own lives. 416\tildes We want a more diverse nature, a more nuanced environment 501\tildes and a richer cultural environment. 502\tildes We want a world where children do well to live. 706}

\section{Sample of statements labelled by the ensemble model}
\label{sec:statement-examples}

In this section, we provide 10 random statements from the UK Hansard for five random MARPOR labels assigned by our consensus ensemble model.

\begin{itemize}
    \small

    \item[201] \enquote{Freedom and Human Rights}
        \begin{itemize}
            \item We hear about the freedom and liberty of the individual yet every so often we see on the Order Paper another of these county council Bills or something of the sort that includes this requirement to give prior notice of processions and demonstrations.
            \item At the very least, it should be an offence to impersonate another person for the purpose of obtaining compulsory access to personal information.
            \item My right hon.\ and hon.\ Friends believe that the civil rights of the citizen come first and foremost.
            \item He had come to similar conclusions over 10 years ago on the same basis — that Parliament could no longer safeguard the liberties of the individual.
            \item — to unconditionally release Nelson Mandela and the other political prisoners
            \item As from 11 November this year, individuals will have the right to demand access to any data held about them on police computer systems and, where appropriate, to have such data corrected or erased.
            \item That, apparently, is what the Prime Minister means by freedom of choice.
            \item The applicant is not told whether information about him or her is held on computer.
            \item Clause 2(1) is most important as it balances the competing interests of freedom of information with the protection of the individual's privacy.
            \item It would be an offence for those responsible for the operation of the police national computer wrongly to disclose such personal information.
        \end{itemize}
    
    \item[202] \enquote{Democracy}
        \begin{itemize}
            \item Will not that be the right time to enter into new discussions?
            \item It also requires us to reassess, as a House, the control that we believe we should exercise on behalf of the people, of the means that we use to protect them.
            \item Let us not be kidded — democracy affects local government.
            \item The Minister who piloted through the Elections (Northern Ireland) Act 1985 — that unwanted piece of legislation — will be well aware that any attempt to filter and vet electors when they present themselves at the entrance to the polling station is illegal under that Act.
            \item Its chairman, John Hosking, and others have taken a considerable interest in the subject.
            \item Will the Leader of the House give us his views on the prospects for a debate on an issue affecting democratic debate in the House?
            \item Is it in order for a group such as the Amalgamated Engineering Union parliamentary Labour group to be a sponsor of a Bill in the House, because it must surely include Members of the other place as well?
            \item As Winston Churchill said, a democracy is an imperfect form of government.
            \item The more one studies that view, however, the more ineffective a weapon it has proved to be for Oppositions over the past 30 years.
            \item Therefore, I shall be as helpful as I can during the Committee stage, provided that Ministers participate fully in the process.
        \end{itemize}

    \item[601] \enquote{National Way of Life: Positive}
        \begin{itemize}
            \item A further consequence of the contradiction between the Government's budgetary and monetary policies is that we shall increase the attractiveness of the United Kingdom as a haven for the world's footloose funds.
            \item Thereby they will lift a burden from the backs of the British people.
            \item  Should it fail, we must use our best endeavours both before and after independence to ensure that nothing disrupts that country.
            \item To do that, they had to have their own citizenship.
            \item They lit bonfires in Marlborough, they had cream teas in Ramsbury, they had special children's fetes in Great Bedwyn and smaller fetes in Little Bedwyn.
            \item As hon.\ Members know, this is Derby day.
            \item Subject to the same safeguards, I believe that the existing law should be extended to provide the same protection for Her Majesty the Queen and the royal family as is now available to foreign embassies and diplomats.
            \item I am convinced that, by those standards, Britain could do better.
            \item I believe that they see themselves more as Londoners now than they did even 18 months ago.
            \item Is it not true that even if they all arrived tomorrow morning, that would still represent only 3 per cent. of the British birth rate and there would still be a net outflow of emigrants from this country?
        \end{itemize}

    \item[603] \enquote{Traditional Morality: Positive}
        \begin{itemize}
            \item We are told that the income tax reduction for the average family is 75p–80p.
            \item Does she realise that any delay will mean that five times the number of babies born in that group will either be born either dead or with a severe handicap?
            \item According to Government figures, 25,000 people who are unemployed and registered for work are unmarried childless couples living together as man and wife.
            \item They are brought out at births, deaths and funerals and, when I visit the Sikh temple in my constituency, they are offered as hospitality and a welcome to worship.
            \item It could be argued — this is why the previous Labour Government backed down on proposals which did not go as far as the present ones — that it is more likely that at the age of 60 family commitments will have decreased.
            \item I have listened carefully to the hon. Gentleman's speech in which he has ranged widely from the Old Testament to the Rocky mountains and back to confessions.
            \item When I was a little boy I was told that I had to work twice as hard as everybody else because I did not have a father.
            \item The old system undoubtedly constituted a tax on marriage in exactly the same way as the former allowance of double tax relief on mortgages for unmarried persons was a tax on marriage.
            \item My husband agreed to have another baby and now I am six months pregnant and we are both overjoyed.
            \item He should stop believing as gospel everything that he reads in the newspapers.
        \end{itemize}

    \item[305] \enquote{Political authority}
        \begin{itemize}
            \item We know the problems, as we have said many times in this House.
            \item That is true.
            \item I was delighted to say the same to you in a similar debate at almost exactly the same time last year.
            \item I am grateful for that reply.
            \item I shall come to the Conservative manifesto.
            \item He is quite right.
            \item Perhaps you can help me by saying whether it is in order to listen to a point of order raised by Liberal Members, all of whom have been absent until 45 minutes ago, who have come into the debate just recently and seem to be voting —
            \item He brought a deputation to my Department last Thursday, and I was extremely impressed by the responsible and well argued approach adopted by the councillors and officials whom I met and by the way that the case had been prepared in some documents which I found compelling reading.
            \item The Minister looks askance at that comment, but he is the only one who has held office in that Department for four years.
            \item I realise that it has been a long evening for Conservative Members and that a large number are being forced to stay here in case the Opposition require a vote to be held later tonight.
        \end{itemize}
\end{itemize}

\section{Hansard UK statistics}
\label{sec:stats}

\begin{figure*}[t]
    \centering
    \includegraphics[width=\linewidth]{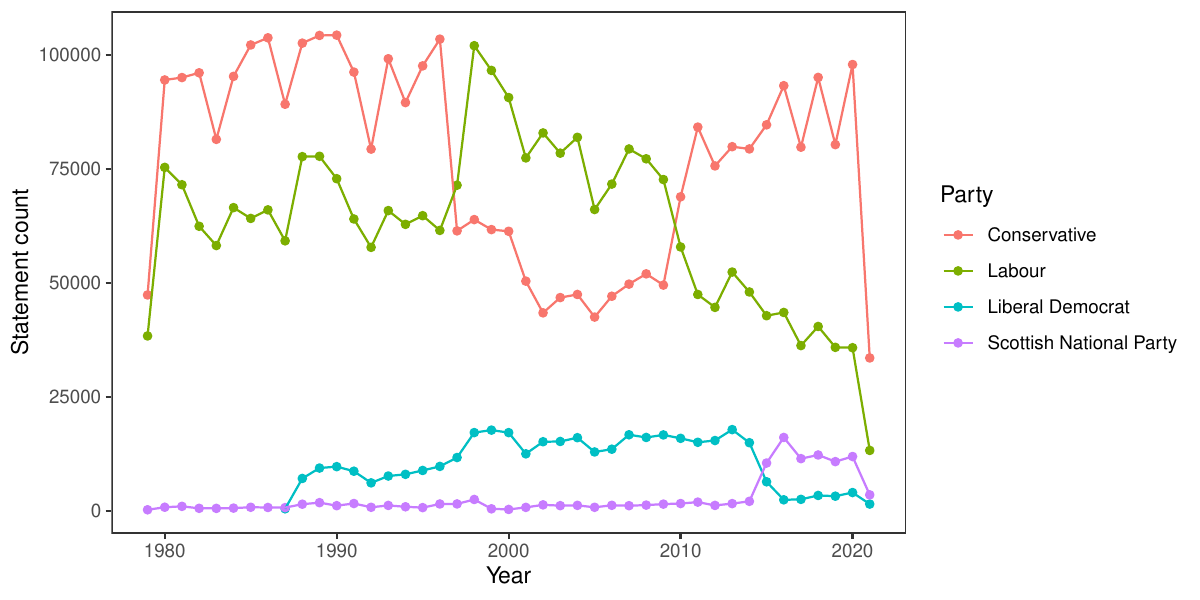}
    \caption{Yearly speech counts of four major UK parties recorded in Hansard over the last four decades.}
    \label{fig:hansard-uk-statements}
\end{figure*}

Statistics of the number of statements made by member of the four major parties in the House of Commons
are shown in Figure~\ref{fig:hansard-uk-statements}.

\section{RILE scores of major UK parties}
\label{sec:uk-riles}

\begin{figure*}
    \centering
    \includegraphics[width=\linewidth]{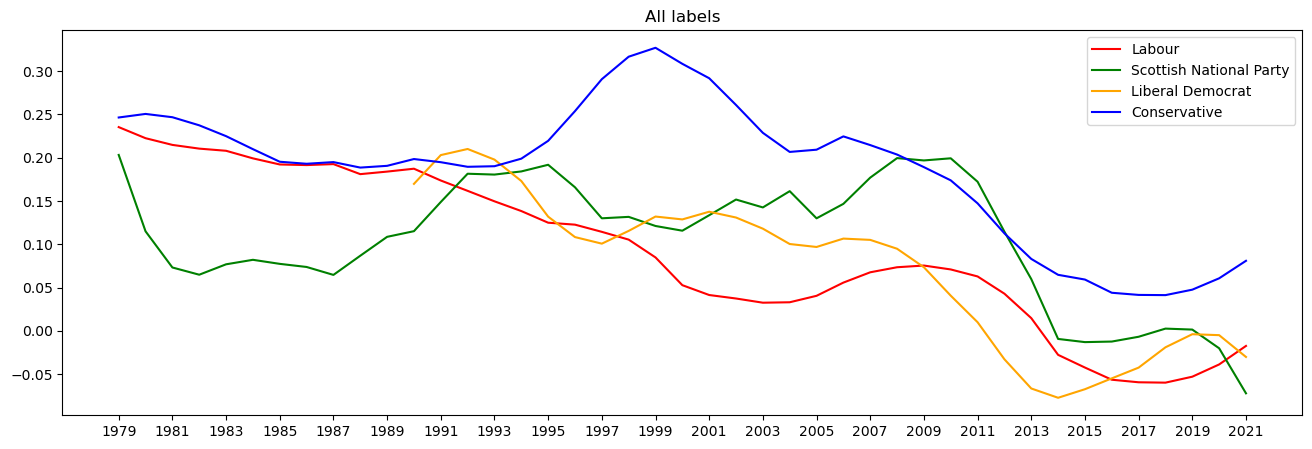}
    \caption{RILE scores for four major UK parties computed based on the House of Commons speeches by their members.}
    \label{fig:hansard-uk-riles}
\end{figure*}

\begin{figure*}
    \centering
    \includegraphics[width=\linewidth]{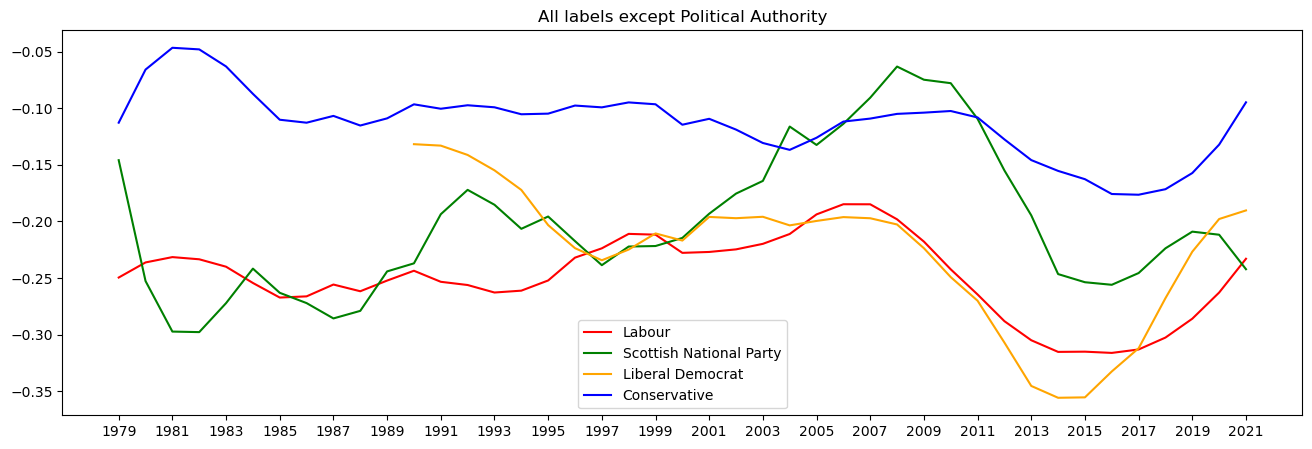}
    \caption{RILE scores for four major UK parties computed based on the House of Commons speeches by their members,
    with label 305, \enquote{Political authority}, excluded from the estimation.}
    \label{fig:hansard-uk-riles-no-305}
\end{figure*}

RILE scores computed on all available data from the UK Hansard are shown in Figure~\ref{fig:hansard-uk-riles} (all labels)
and Figure~\ref{fig:hansard-uk-riles-no-305} (all labels except 305, \enquote{Political authority}, which is equally 
overpredicted for all parties and does not influence their mutual differences but shifts all RILE scores to the right).


Conservatives are consistently portrayed as the most right-wing party, with SNP briefly overtaking them in the
run-up to the referendum on Scottish independence, which took place in 2014. After the independence was rejected
by the voters, SNP returned to its other traditional focus on social-welfare issues.

\section{Trajectories of Australian parties}
\label{sec:hansard-au-analysis}

\begin{figure*}[t]
    \centering
    \includegraphics[width=\linewidth]{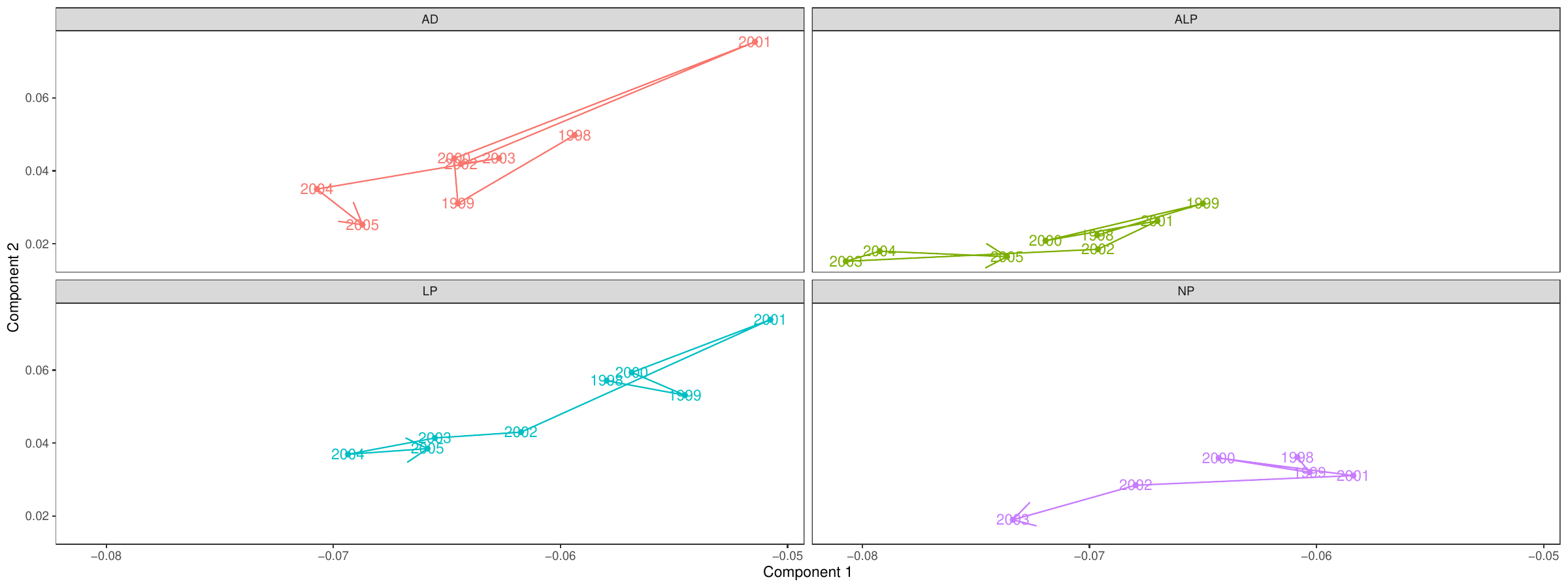}
    \caption{Political trajectories of four major Australian parties traced by projecting yearly salience vectors of CMP categories in their speeches in Parliament (both House of Representatives and Senate) using non-negative matrix factorization and the original CMP data as the training set. AD: Australian Democrats; ALP: Australian Labour Party; LP: Liberal Party; NP: National Party.}
    \label{fig:nmf-trajectories-au}
\end{figure*}

Original XML files published by the Australian Parliament and provided by \citet{sherrat2019au}
were used to extract the statements for analysis. Only the subset from 1998 till 2005 was analyzed.
See \citet{katz2023hansard} for a more up-to-date dataset.
The results of the application of NMF-based analysis to the data are shown in Figure~\ref{fig:nmf-trajectories-au}.

\end{document}